\pdfoutput=1

\documentclass[11pt]{article}

\usepackage{multirow}
\usepackage{multicol}
\usepackage{booktabs}
\usepackage{hyperref}
\usepackage[dvipsnames]{xcolor}
\usepackage{times}
\usepackage{booktabs} 
\usepackage{latexsym}
\usepackage{subcaption}
\usepackage{graphicx}

\usepackage{amsmath}
\usepackage{amssymb}
\usepackage{mathtools}
\usepackage{amsthm}
\usepackage{caption}
\usepackage[capitalize,noabbrev]{cleveref}
\usepackage{amsfonts}
\usepackage{microtype}
\usepackage{svg}
\usepackage{soul}

\usepackage{bm}
\usepackage[ruled,vlined]{algorithm2e}
\usepackage{verbatim}

\usepackage[]{EMNLP2022}

\usepackage{times}
\usepackage{latexsym}

\usepackage[T1]{fontenc}

\usepackage[utf8]{inputenc}

\usepackage{microtype}

\usepackage{inconsolata}

\title{Candidate Soups: Fusing Candidate Results Improves Translation Quality for Non-Autoregressive Translation}

\author{
Huanran Zheng$^{1}$, Wei Zhu$^{1}$, Pengfei Wang$^{1}$, Xiaoling Wang$^{1}$\\
    $^{1}$East China Normal University, Shanghai, China\\
  \texttt{\{hrzheng,wzhu,pfwang\}@stu.ecnu.edu.cn} \\  
  \texttt{\{xlwang\}@cs.ecnu.edu.cn}
 }

\begin{document}
\maketitle
\begin{abstract}
Non-autoregressive translation (NAT) model achieves a much faster inference speed than the autoregressive translation (AT) model because it can simultaneously predict all tokens during inference. However, its translation quality suffers from degradation compared to AT. And existing NAT methods only focus on improving the NAT model's performance but do not fully utilize it. In this paper, we propose a simple but effective method called “Candidate Soups,” which can obtain high-quality translations while maintaining the inference speed of NAT models. Unlike previous approaches that pick the individual result and discard the remainders, Candidate Soups (CDS) can fully use the valuable information in the different candidate translations through model uncertainty. Extensive experiments on two benchmarks (WMT’14 EN–DE and WMT’16 EN–RO) demonstrate the effectiveness and generality of our proposed method, which can significantly improve the translation quality of various base models. More notably, our best variant outperforms the AT model on three translation tasks with \textbf{7.6$\times$} speedup.\footnote{Our code
is released at \url{https://github.com/boom-R123/Candidate_Soups}.}
\end{abstract}

\section{Introduction}
\label{sec:intro}

Autoregressive translation (AT) models based on Transformer \cite{vaswani2017attention,So2019TheET,sun-etal-2022-simple,zhu-etal-2021-gaml}, where each generation step depends on the previously generated tokens, achieve state-of-the-art (SOTA) performance on most datasets for machine translation tasks. AT model can better model the process of translation generation but leads to a massive limitation of its inference speed. 

\begin{figure}[t]
    \centering
    \includegraphics[width=0.42\textwidth]{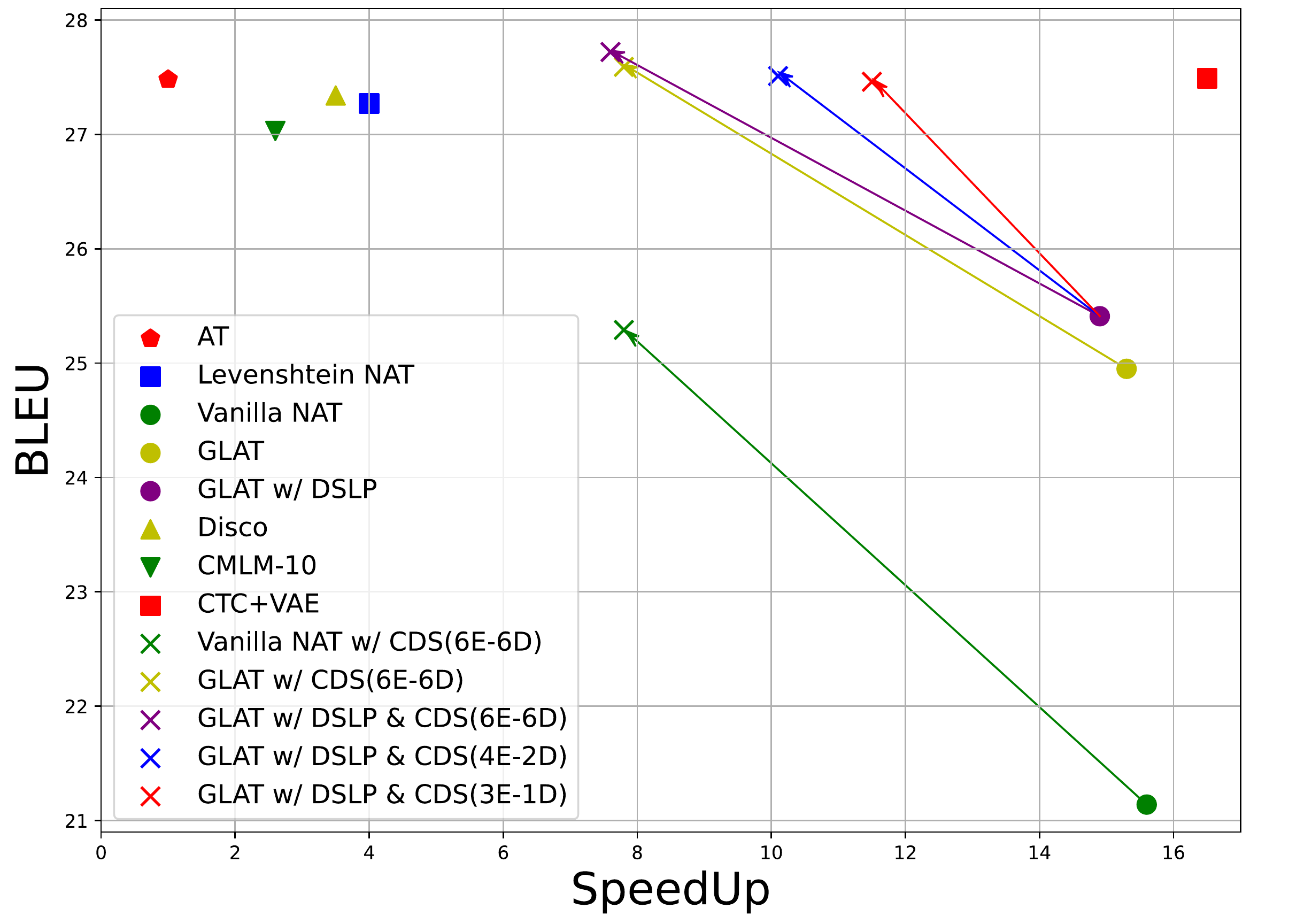}
    \caption{Efficiency (Speedup) and Translation quality (BLEU) of NAT models in the WMT’14 EN-DE translation dataset. A cross “×” represents our Candidate Soups (CDS) variants. Its base model is shown in the shape “$\bullet$”, and its correspondence is represented by an arrow. CDS (mE-nD) refers to the AT model for re-scoring that has m encoder layers and n decoder layers.}
    \label{fig:BLEU_speed}
\end{figure}

Therefore, the non-autoregressive translation (NAT)~\citep{gu2018non} model is proposed, which is 15.6 times faster than AT model. NAT assumes that the generated tokens are conditionally independent given the source sentence, so the translation can be generated in parallel, significantly improving its inference speed compared to AT. However, due to the strong independence assumption, the ability of the NAT model modeling sequence generation is weakened. So NAT model usually has multimodality problem~\citep{gu2018non} in the inference process, resulting in its performance worse than AT models.

\begin{figure*}[t]
    \centering
    \normalsize
    \renewcommand{\arraystretch}{1.03}
    \begin{tabular}{ll}
        \toprule
        Src & Die Beschaffung des erforderlichen Personalausweises kostet oft über hundert Dollar. \\
        \hline
        Candidate 1 & It often costs over a hundred dollars to obtain the \textcolor{red}{require} identity card .\\
        Candidate 2 & It often \textcolor{red}{cost} over a hundred dollars to obtain the required identity card . \\
        \hline
        NPD & It often \textcolor{red}{cost}  over a hundred dollars to obtain the required identity card . \\
        CDS & It often costs over a hundred dollars to obtain the required identity card .\\
        \bottomrule
    \end{tabular}
     \caption{Comparison between NPD and Candidate Soups (CDS).  \textcolor{red}{Red fonts} represent mistranslated tokens. Compared with NPD, Candidate Soups does not preserve the erroneous parts in the candidate results.}
    \label{fig:Comparison of NPD and Candidate Soups}
\end{figure*}

Several methods have been proposed to alleviate the multimodality problem and improve the performance of the NAT model, such as the iteration-based NAT model~\cite{ghazvininejad2019mask,gu2019levenshtein,kasai2020non} and the semi-autoregressive translation model~\cite{Wang2018SemiAutoregressiveNM,ran2020learning}.

Most of the previous methods are modified from the model's perspective, either modifying the structure of the model~\cite{shu2020latent,huang2021non,Zhu2021AutoTransAT} or modifying the training method of the model~\cite{du2021order,qian2020glancing}. Different from the previous methods, in this paper, we propose a simple but effective method: Candidate Soups, which can significantly improve the translation quality without any modification to the model.
Moreover, Candidate Soups is a general approach that can be used by any NAT model that can generate multiple candidate results, such as Vanilla NAT~\cite{gu2018non}, GLAT~\cite{qian2020glancing}, etc.

The conventional recipe for maximizing translation quality through candidate results is noisy parallel decoding (NPD)~\cite{gu2018non}, which regards each candidate translation as an independent individual and ultimately only selects one of them as the final result and discards others. Therefore NPD can not utilize the valuable information in all the candidate translations. For example, there are a total of two candidate translations. The first candidate translation has the wrongly translated word in the second half, and the second candidate translation has the wrongly translated word in the first half. Using the NPD algorithm, in this case, can not get the correct translations (Figure \ref{fig:Comparison of NPD and Candidate Soups}). 

However, Candidate Soups will effectively use the valuable information of all the candidate translations to fuse the different candidate results and obtain a higher-quality translation (Figure \ref{fig:Comparison of NPD and Candidate Soups}). Specifically, Candidate Soups first finds the common subsequence among all candidate results. Based on the uncertainty of the model, we consider the common subsequence to be the most confident part of the model's predictions, so we make it part of the final translation and use it to align the candidate results. For the remaining parts, we select the part with the highest average log-probability among all candidate results to add to the final translation. Candidate Soups can be regarded as an implicit model ensemble method, which generates multiple different results by introducing uncertainty in the inference process, and further enhances the translation quality by making full use of the information of multiple results.

We conduct extensive experiments in two datasets commonly used in machine translation, WMT’14 EN–DE and WMT’16 EN–RO. The results demonstrate that our proposed method can significantly improve the base models' translation quality on different tasks while maintaining the fast inference speed of the NAT model. Remarkably, our best variant achieves better performance than the AT teacher model on three translation tasks with  \textbf{7.6$\times$} speedup. Figure \ref{fig:BLEU_speed} demonstrates the quality-speed trade-off compared with AT and recent NAT models. And relevant background knowledge is introduced in Appendix \ref{appendix:background}.


\section{Related Work}
Since the NAT model was proposed, it has attracted the attention of many researchers due to its superior inference speed. However, its translation quality suffers from degradation compared to AT model. Therefore various methods have been proposed to bridge the performance gap between NAT and AT model.

Some researchers constrain the distribution of NAT model outputs by introducing various latent variables~\cite{gu2018non,shu2020latent,ran2021guiding}. Through latent variables, the diversity of the NAT model output space can be significantly reduced so that the model can better handle the dependencies between output words and alleviate the multimodality problem. Such methods can usually maintain the efficient inference speed of the NAT model, but the improvement in translation quality is relatively small.

Some other researchers have proposed iterative decoding methods~\cite{ghazvininejad2019mask,gu2019levenshtein,kasai2020non}, which continuously optimize the model's output by introducing more information in the iterative process. For example, \citet{ghazvininejad2019mask} mask the partial token of the previous output result and then use it as the input of the decoder for the next round of iteration. Although such models can achieve high performance, multiple iterations can also significantly affect the inference speed of the NAT model.

\begin{figure}[htbp]
	\centering
	\begin{subfigure}{0.9\linewidth}
		\centering
		\includegraphics[width=0.9\linewidth]{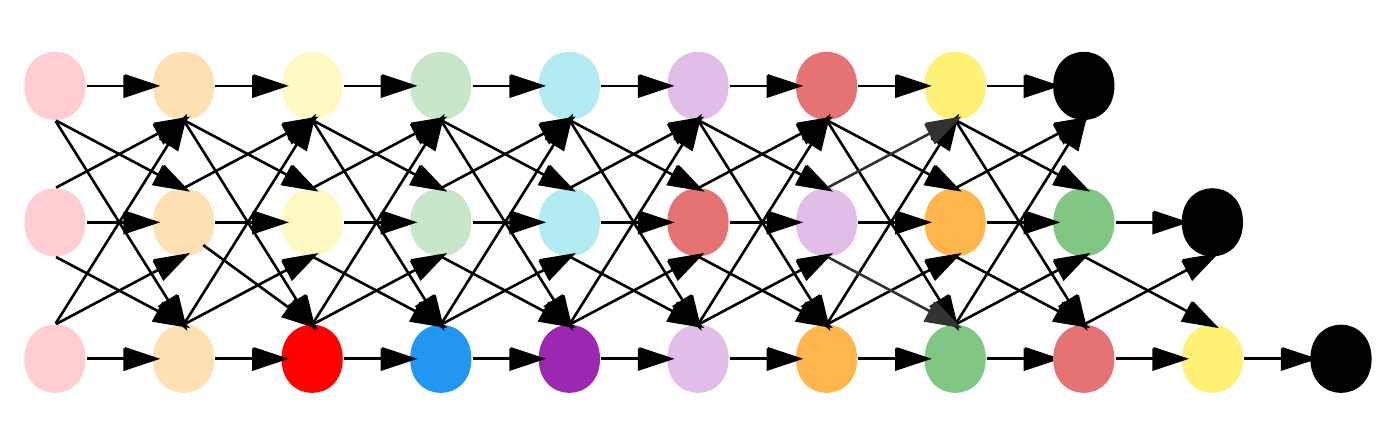}
		\caption{Initial Lattice}
		\label{Problem_definition_a}
	\end{subfigure}
	\begin{subfigure}{0.9\linewidth}
		\centering
		\includegraphics[width=0.9\linewidth]{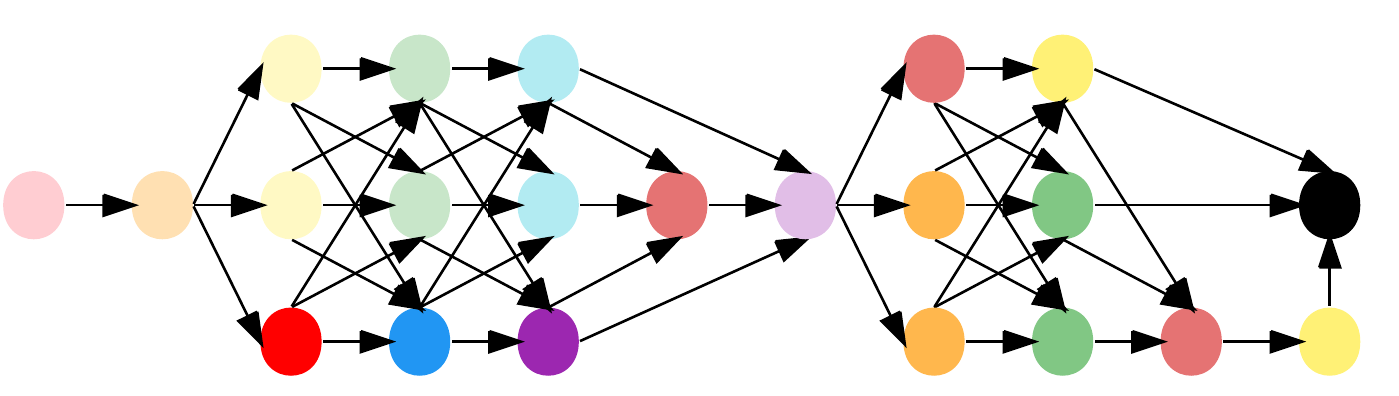}
		\caption{Simplified Lattice}
		\label{Problem_definition_b}
	\end{subfigure}
	
	\begin{subfigure}{0.9\linewidth}
		\centering
		\includegraphics[width=0.9\linewidth]{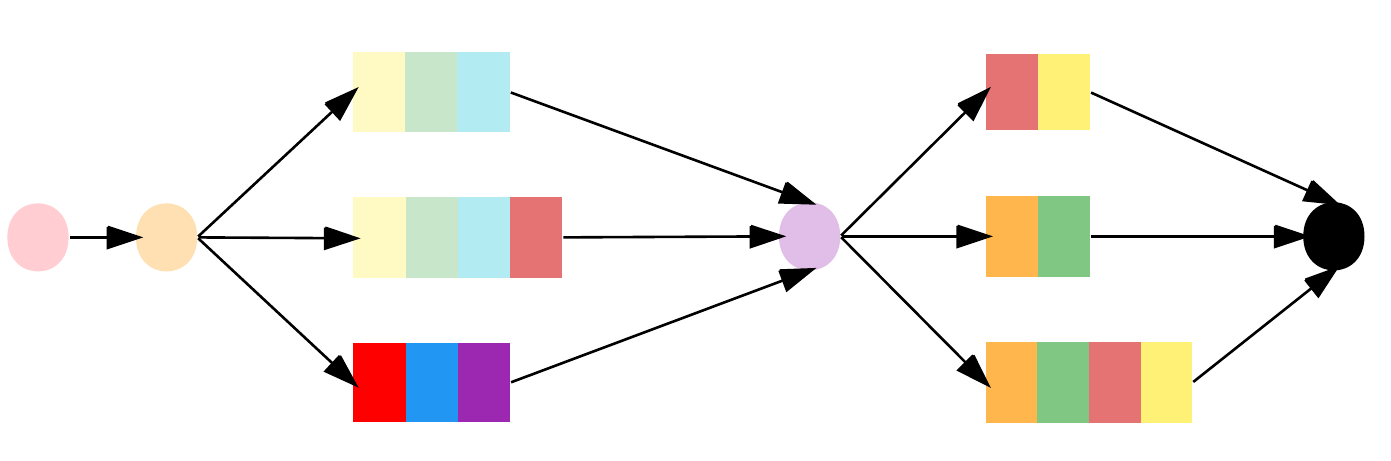}
		\caption{Final Lattice}
		\label{Problem_definition_c}
	\end{subfigure}
\caption{Definition and simplification of translation search problem. Assume there are three candidate translations. Nodes with different colors represent different tokens.
(\subref{Problem_definition_a}) Initial search space.
(\subref{Problem_definition_b}) Simplified search space using the common subsequence.
(\subref{Problem_definition_c}) Further simplified search space after node fusion.
}
\label{fig:Problem_definition}
\end{figure}

Recently, \citet{qian2020glancing} borrowed ideas from curriculum learning and proposed a novel way to train NAT models, which let the model starts from learning the generation of sequence fragments and gradually moving to whole sequences. \citet{huang2021non} proposes to predict the result at each decoder layer and input it to the next layer together with the output of the current decoder layer to modify the result of the subsequent prediction.

The above methods are all improvements from the model perspective, and their purpose is to allow the model to generate higher quality translations. Unlike previous work, Candidate Soups wants to explore how to make the most of an existing model, and it can be applied to all NAT models that can generate multiple candidate results.

\section{Candidate Soups}
This section describes the details of the proposed method in the paper. We first show the problem definition and the general idea of Candidate Soups in Section \ref{Problem definition}, then introduce implementation details of the Candidate Soups in Section \ref{Implementation}, followed by the example in Section \ref{Example}.

\subsection{Problem Definition}
\label{Problem definition}
By introducing uncertainty into the NAT model, we can get a list of candidate results $\bm{R}={\left[R_0, \ldots, R_k\right]}$, and each candidate result may have correctly and incorrectly translated parts that do not completely overlap. Thus, our goal is to find the optimal combination in $\bm{R}$ , which has the highest average log-probability re-scored by an AT model. Because the word order in the original translations must be kept, we first use $\bm{R}$ to build a Lattice (Figure \ref{Problem_definition_a}). Each node represents a token, and each edge represents the change of average log-probability caused by adding the next token. So each path from the beginning node \textbf{[BOS]} to the end node \textbf{[EOS]} represents a possible translation. Therefore, our goal is to find the best path which has the highest average log-probability in this Lattice. 

\begin{algorithm}[t]
\footnotesize
\caption{Candidate Soups}
\textbf{Input:} A list of candidate results $\bm{R}={\left[R_0, \ldots, R_k\right]}$ and the corresponding log-probability score sequence list $\bm{S}={\left[S_0, \ldots, S_k\right]}$ \\
Result = [] \\
$\bm{R}$, $\bm{S}$ = Remove$\_$duplicates($\bm{R}$, $\bm{S}$)\\
Initialize I = [$i_0$ = 0, \ldots, $i_k$=0] \\
Get candidate results length list L = [$l_0$, \ldots, $l_k$]\\
\While{I $<$ L}{
\eIf{all item in $\bm R$[I] is equal}{
    Result.append($R_0$[$i_0$])\\
    I = I + 1\\
}
{
$\bm H$ = $\bm R$[I] \\
I$^*$ = I + 1\\
\tcp{H is used to record tokens going from I to I$^*$}

    \While{I$^*$ $<$ L}{
         Add $\bm R$[I$^*$] to $\bm H$ \\
         \If{Exist $\hat{I}$ meet all item in $\bm H$[$\hat{I}$] is equal}{
            I$^*$ = $\hat{I}$ \\
            break
         }
         I$^*$ = I$^*$ + 1
    }
    T = [$S_j$[$i_j$-1:$i^*_j$+1].mean() for j from 0 to k] \\
    t = T.argmax() \\
    Result.append($R_t$[$i_t$:$i^*_t$]) \\
    I = I$^*$
}

}
return Result
\label{alg:Candidate Soups}
\end{algorithm}

\begin{figure*}[th]
    \centering
    \renewcommand{\arraystretch}{1.09} 
    \resizebox{0.95\linewidth}{!}{\begin{tabular}{lll} 
        \toprule
        \multicolumn{2}{c}{Src} & Die republikanischen Behörden beeilten sich , diese Praxis auf andere Staaten auszudehnen . \\
        \hline
        \multirow{3}*{t=0}
        & Candidate 1 & The Republican authorities were quick extend to other States . \\
        & Candidate 2 & The Republican authorities were quick to extend this practice States . \\
        & Candidate 3 & The Republican and the authority extend this practice to other States . \\
        \hline
        \multirow{3}*{t=1}
        & Candidate 1 & \hl{The Republican} authorities were quick extend to other States . \\
        & Candidate 2 & \hl{The Republican} authorities were quick to extend this practice States . \\
        & Candidate 3 & \hl{The Republican} and the authority extend this practice to other States . \\
        \hline
        \multirow{3}*{t=2}
        & Candidate 1 & \hl{The Republican} \textcolor{red}{authorities were quick} extend to other States . \\
        & Candidate 2 & \hl{The Republican authorities were quick to} extend this practice States . \\
        & Candidate 3 & \hl{The Republican} \textcolor{red}{and the authority} extend this practice to other States . \\
        \hline
        \multirow{3}*{t=3}
        & Candidate 1 & \hl{The Republican} \textcolor{red}{authorities were quick} \hl{extend} to other States . \\
        & Candidate 2 & \hl{The Republican authorities were quick to extend} this practice States . \\
        & Candidate 3 & \hl{The Republican} \textcolor{red}{and the authority} \hl{extend} this practice to other States . \\
        \hline
        \multirow{3}*{t=4}
        & Candidate 1 & \hl{The Republican} \textcolor{red}{authorities were quick} \hl{extend} \textcolor{red}{to other} States . \\
        & Candidate 2 & \hl{The Republican authorities were quick to extend} \textcolor{red}{this practice} States . \\
        & Candidate 3 & \hl{The Republican} \textcolor{red}{and the authority} \hl{extend this practice to other} States . \\
        \hline
        \multirow{3}*{t=5}
        & Candidate 1 & \hl{The Republican} \textcolor{red}{authorities were quick} \hl{extend} \textcolor{red}{to other} \hl{States .} \\
        & Candidate 2 & \hl{The Republican authorities were quick to extend} \textcolor{red}{this practice} \hl{States .} \\
        & Candidate 3 & \hl{The Republican} \textcolor{red}{and the authority} \hl{extend this practice to other States .}  \\
        \hline
        \multicolumn{2}{c}{Final Result} & The Republican authorities were quick to extend this practice to other States . \\
        \bottomrule
    \end{tabular}}
     \caption{An example from the WMT’14 DE-EN validation set illustrates how Candidate Soups generates high-quality final translations from candidate results. The \hl{highlighted tokens} represent the tokens added to the final translation, and the \textcolor{red}{red tokens} represent the discarded tokens.}
    \label{fig:Example}
\end{figure*}

However, because the initial Lattice contains too many paths, we cannot calculate the values of all edges. Furthermore, due to the dislocation caused by the different lengths of the candidate results, most of the paths in the initial Lattice have word order errors, such as edges between the same tokens (Figure \ref{Problem_definition_a}). So we simplified and aligned the original Lattice. Specifically, after introducing the length uncertainty, we consider tokens that appear in all candidate results with the same word order as certain components~\cite{Gal2016DropoutAA}. Thus, we find the common subsequence of all candidate results and directly use it as part of the final translation (Figure \ref{Problem_definition_b}). This way, the alignment of candidate translations can be achieved, and the original complex Lattice can be simplified into the connection of multiple simple Lattices.

For the remaining simple Lattice, the cost of calculating each edge value is still unbearable. Therefore, we fuse the nodes belonging to the same candidate result in each Lattice into a single node (Figure \ref{Problem_definition_c}) and ignore the influence of previous Lattice results on subsequent Lattice. In this way, we can quickly calculate each edge value by using the AT model to re-score each candidate translation. Then we only need to calculate the best path in each simple Lattice and obtain the final translation through a simple greedy algorithm.

\subsection{Implementation}
\label{Implementation}
Algorithm \ref{alg:Candidate Soups} lists the process of Candidate Soups. We generate the final translation while looking for the common subsequence. 

First, for the input candidate results set $\bm R$ and the corresponding log-probability score set $\bm S$, we will remove the adjacent repeated tokens and their corresponding scores for each sentence. Then we initialize a pointer set $I$ that each pointer points to position 0 for each candidate translations and use these pointers to traverse simultaneously. If all the current pointers point to the same token, the token is added to the final translation, and all pointers are moved one step to the right. Otherwise, Candidate Soups will look for the next sequence of pointers $I^*$ that satisfies the above conditions and move all pointers there. At the same time, the segment with the highest average log-probability score among all segments generated by the pointer traverse from $I$ to $I^*$ is added to the final translation.

Experimental results show that Candidate Soups can significantly improve final translation quality, requiring only 3 to 7 candidate translations. Moreover, the time required by Candidate Soups is almost negligible compared to the inference time of the NAT model.

\subsection{Example}
\label{Example}

Figure \ref{fig:Example} shows how Candidate Soups fuses the valuable information in each candidate translation to generate high-quality translations during the traversal process. 

First, the NAT model predicts three candidate translations for the input sentence by introducing different lengths (t = 0). Afterward, through the traversal of the pointers, Candidate Soups found that the first two tokens (“The Republican”) in the candidate results were the same, so they were added to the final translation (t = 1). When there is a disagreement between candidate results, Candidate Soups will find the next token (“extend”) that all candidate translations predict in common and get three different segments (“authorities were quick,” “authorities were quick to,” and “and the authority”). Then Candidate Soups will select the candidate segment with the highest average log-probability scores (“authorities were quick to”) and add it to the final translation (t = 2). Similarly, in the subsequent traversal process, if the tokens are predicted jointly by all candidate results, Candidate Soups will add them to the final translation (t = 3, t = 5). Otherwise, Candidate Soups will select the tokens segment with the highest log-probability score to join the final translation (t = 4). Ultimately, we get higher-quality translations by combining all valuable information in the candidate results. 

From this example, we can find that only selecting an independent candidate result as the final translation is not effective enough for the NAT model. Because different lengths introduce uncertainty into the NAT model, there is diversity among candidate translations, but the previous methods do not take advantage of this. Proudly, through the certainty and confidence of the NAT model's output, Candidate Soups makes full use of the candidate results, takes the essence and removes the dross, and further improves the final translation quality without affecting the inference speed.

\section{Experiments}
\label{sec:experiment}
In this section, we first introduce the settings of our
experiments in Section \ref{subsec:setup}, then report the main results in Section \ref{main_results}. Ablation experiments and analysis are presented in Section \ref{analysis}.

\begin{table*}[t]
\vspace{-.1cm}
\centering
\renewcommand{\arraystretch}{0.9}
\resizebox{.7\linewidth}{!}
{
    \begin{tabular}{rl|cc|cc|cl} 
    \toprule
           \multirow{2}{*}{Row\#}   &\multirow{2}{*}{Model}& \multicolumn{2}{c|}{WMT'14} & \multicolumn{2}{c|}{WMT'16}  & \multirow{2}{*}{Speedup} \\
             &    & EN--DE        & DE--EN       & EN--RO           & RO--EN    &                          &                           \\ 
             \midrule
1  & Transformer (teacher)         & 27.48        & 31.21       & 33.70           & 34.05              &        1$\times$                     \\ 
\midrule
2 & Vanilla NAT        & 21.14        & 24.65       &   29.14         & 28.93                  &         15.6$\times$                   \\
3 & ~~~~ w/ NPD & 23.44        & 27.25       &  31.70     & 30.93              &    7.9$\times$    \\
4 & ~~~~ w/ Candidate Soups & \textbf{25.29}        & \textbf{28.50}       &  \textbf{32.75}     & \textbf{32.22}              &    7.8$\times$                     \\ 
\midrule
5 & CMLM$_1$     & 19.44        &  23.16      &   27.61         & 28.33       &    15.6$\times$     \\ 
6 & ~~~~ w/ NPD  & 21.72       &     26.00  &   30.54       & 31.34        &   7.9$\times$                         \\ 
7 & ~~~~ w/ Candidate Soups & \textbf{23.51}     &     \textbf{27.67}  &   \textbf{32.13}        & \textbf{32.70}        &   7.8$\times$                         \\ 
8 & CMLM$_5$     & 26.37        &  30.05      &   32.30        & 30.73       &    5.3$\times$     \\ 
9 & ~~~~ w/ NPD  & 26.94       &     30.68  &   33.07        & 33.68        &   4.4$\times$                         \\ 
10 & ~~~~ w/ Candidate Soups  & \textbf{27.80}        &     \textbf{31.21}  &   \textbf{33.42}         &  \textbf{34.06}        &   4.4$\times$                         \\ 
\midrule
11 & GLAT     & 24.95        &  28.80      &   31.29         & 31.93       &    15.6$\times$     \\ 
12 & ~~~~ w/ NPD  & 26.19       &     30.64  &   32.46        & 33.38        &   7.9$\times$                         \\ 
13 & ~~~~ w/ Candidate Soups  & \textbf{27.59}        &     \textbf{30.95}  &   \textbf{33.22}         & \textbf{33.73}        &   7.8$\times$                         \\ 
\midrule
14 & GLAT \& DSLP       & 25.41        & 29.28       &   32.32         & 32.37               &       14.8$\times$     \\
15 & ~~~~ w/ NPD    & 26.68        & 30.69       &   33.32         & 33.67          &  7.7$\times$                         \\ 
16 & ~~~~ w/ Candidate Soups    & \textbf{27.72}        & \textbf{30.98}       &   \textbf{33.71}         & \textbf{34.11}          &  7.6$\times$                         \\ 
\midrule

& Average Improvement   &  2.92    &  2.67        &  2.51   &  2.91   & --       \\ 
\bottomrule

\end{tabular}
}
\vspace{-.1cm}
\caption{Applying Candidate Soups to four different base NAT models, which shows the generality of our algorithm. Translation quality is evaluated in BLEU. Speedup is relative to the AT teacher. All results are achieved by us, except Transformer (teacher), which is obtained from \citet{huang2021non}. CMLM$_k$ refers to k iterations of progressive generation. Here, we consider k = 1 and k = 5. Both NPD and Candidate Soups use the AT model to re-score, and the number of candidate results is 5.}
\label{tab:add2base}
\vspace{-.1cm}
\end{table*}

\subsection{Experimental Setup}
\label{subsec:setup}
\paragraph{Dataset and Evaluation}
We evaluate our method on the two most recognized machine translation benchmarks: WMT’14 English–German (4.0M sentence pairs)\footnote{\url{https://www.statmt.org/wmt14.}} and WMT’16 English–Romanian (610K pairs)\footnote{\url{https://www.statmt.org/wmt16.}}. We use BLEU~\cite{papineni2002bleu} to evaluate the translation quality. And the beam size is set to 5 for AT model during inference. Moreover, for a fair comparison, we obtain the \citet{huang2021non} open-source corpus whose tokenization and vocabulary are the same as previous work: \citet{Zhou2020UnderstandingKD} for WMT’14 EN–DE which contains 39.8k subwords, and \citet{Lee2018DeterministicNN} for WMT’16 EN–RO which  contains 34.6k subwords. Both the implementation and evaluation of our method are performed using the open source fairseq\footnote{\url{https://github.com/pytorch/fairseq.}}~\cite{ott-etal-2019-fairseq}.

\paragraph{Knowledge Distillation} 
Using AT model’s output to train the NAT model can significantly improve the performance of the NAT model. Following previous work~\cite{gu2018non, Lee2018DeterministicNN, ghazvininejad2019mask}, we also employ sequence-level knowledge distillation for all datasets. All the distillation data we use is open sourced by~\citet{huang2021non}.

\paragraph{Hyperparameters} 
Our model architecture is Transformer-base~\cite{vaswani2017attention}: a 6-layer encoder
and a 6-layer decoder, 8 attention heads per layer, 512 attention modules dimensions, 2048 feedforward modules hidden dimensions. We
adopt the Adam optimizer~\cite{kingma2015adam} with $\beta=(0.9,0.98)$. To train the models, we use a batch size
of 64K tokens, with a maximum 300K updates. For regularization, we use dropout (WMT'14 EN-DE: 0.1, WMT'16 EN-RO: 0.3), 0.01 weight decay and 0.1 label smoothing.

\paragraph{Base Models} 
Our Candidate Soups is a general algorithm that can be applied to various NAT models. Therefore, to evaluate whether our proposed method can perform well on different NAT models, we selected the following four base models:
\begin{enumerate} \setlength{\itemsep}{0pt}
  \item [(1)] Vanilla NAT~\cite{gu2018non}, which predicts length instead of fertility sequence.
  \item [(2)] CMLM~\cite{ghazvininejad2019mask}, whose training strategy follows a masked language model approach similar to BERT~\cite{Devlin2019BERTPO}. And it can perform iterative decoding during inference. We trained one CMLM model for each translation task and respectively iterated decoding once and iterated decoding five times as two baselines.
  \item [(3)] GLAT~\cite{qian2020glancing}, which trains NAT model step-by-step in a curriculum learning manner.
  \item [(4)] GLAT \& DSLP~\cite{huang2021non}, whose decoder layers can get the prediction result of the previous layer. 
\end{enumerate}
The prediction patterns and performance of these base models are quite different, so through them, we can verify whether Candidate Soups can be applied to various NAT models. In the future, we will test Candidate Soups on more NAT models, such as CTC~\cite{libovicky2018end} and CTC+VAE~\cite{gu2020fully}.

\begin{table*}[th!]
\centering
\renewcommand{\arraystretch}{1.01}
\small
\resizebox{0.95\linewidth}{!}{\begin{tabular}{llcccccc}
\toprule
\multicolumn{2}{l}{\multirow{2}{*}{\textbf{Models}}} & 
\multirow{2}{*}{\textbf{Iter.}} &
\multirow{2}{*}{\textbf{Speedup}}&
\multicolumn{2}{c}{\textbf{WMT'14}} & \multicolumn{2}{c}{\textbf{WMT'16}} \\
\multicolumn{2}{c}{} & & & \textbf{EN-DE} & \textbf{DE-EN} & \textbf{EN-RO} & \textbf{RO-EN} \\
\midrule
\multirow{1}{*}{AT}
& Transformer \textit{base} (teacher)& N & 1.0$\times$ &  \bf{27.48}&\bf{31.21}&\bf{33.70}&\bf{34.05} \\
\cmidrule[0.6pt](lr){1-8}
\multirow{9}{*}{Iterative NAT}
& InsT~\cite{stern2019insertion} & $\approx$log N & 4.8$\times$ & 27.41 & - & - & - \\
& CMLM~\cite{ghazvininejad2019mask}$^*$ & 10 & 1.7$\times$ & 27.03 & 30.53 & 33.08 & 33.31 \\
& LevT~\cite{gu2019levenshtein} & Adv. & 4.0$\times$ & 27.27 & - & - & 33.26 \\
& JM-NAT~\cite{guo2020jointly}$^*$ & 10 & 5.7$\times$ & 27.69 & \bf{32.24} & 33.52 & 33.72\\
& DisCO~\cite{kasai2020non}$^*$ & Adv. & 3.5$\times$ & 
27.34 & 31.31 & 33.22 & 33.25  \\
& SMART~\cite{ghazvininejad2020semi}$^*$ & 10 & 1.7$\times$ & 
27.65 & 31.27 & - & -  \\
& Imputer~\cite{saharia2020non}$^*$ & 8& 3.9$\times$ & \bf{28.20} & 31.80 & \bf{34.40} & \bf{34.10} \\
&Multi-Task NAT~\cite{hao2021multi}$^*$&10&1.7$\times$&27.98&31.27&33.80&33.60\\
& RewriteNAT~\cite{geng2021learning}$^*$&Adv.&-&27.83& 31.52& 33.63 &34.09 \\
\cmidrule[0.6pt](lr){1-8}
\multirow{13}{*}{Fully NAT}
& Vanilla NAT~\cite{gu2018non} &  1 & 15.6$\times$ & 17.69 & 21.47 & 27.29 & 29.06 \\
& DCRF~\cite{sun2019fast} & 1 & 10.4$\times$ & 23.44 & 27.22 &  - & -\\
& Flowseq~\cite{ma2019flowseq} & 1 & 1.1 $\times$ & 23.72 &  28.39 & 29.73  & 30.72\\
& ReorderNAT~\cite{ran2020learning} & 1 & 16.1$\times$ & 22.79 & 27.28 & 29.30 & 29.50 \\
& AXE~\cite{ghazvininejad2020aligned}$^*$ & 1 & 15.3$\times$ & 23.53 & 27.90 & 30.75 & 31.54 \\
& ENGINE~\cite{tu-etal-2020-engine} & 1 & 15.3$\times$ &22.15 & - & - & 33.16\\
& Imputer~\cite{saharia2020non}$^*$& 1 & 18.6$\times$  & 25.80 & 28.40 & 32.30 & 31.70 \\
& AlignNART~\cite{song2021alignart} & 1 & 13.4$\times$  &26.40 & 30.40 & 32.50 & 33.10 \\
& OAXE~\cite{du2021order} &1&15.3$\times$&26.10&30.20&32.40&33.30\\
& CTC+VAE~\cite{gu2020fully} & 1 & 16.5$\times$  &\textbf{27.49} & \textbf{31.10}& \textbf{33.79} & \textbf{33.87} \\
& GLAT+NPD~\cite{qian2020glancing} & 1 & 7.9$\times$ & 26.55 & 31.02 & 32.87 & 33.51 \\
& GLAT+DSLP~\cite{huang2021non} & 1 & 14.9$\times$ & 25.69 & 29.90 & 32.36 & 33.06 \\
\cmidrule[0.6pt](lr){1-8}
\multirow{3}{*}{Ours}
& GLAT+DSLP+Candidate Soups (AT 6E-6D) &  1 & 7.6$\times$ &\textbf{27.72} &\textbf{30.98} &\textbf{33.71} &\textbf{34.11}\\
& GLAT+DSLP+Candidate Soups (AT 4E-2D) &  1 & 10.1$\times$ &27.51 & 30.79 & 33.58  & 34.03 \\
& GLAT+DSLP+Candidate Soups (AT 3E-1D) &  1 & 11.5$\times$ &27.46 & 30.69 & 33.65  & 34.01 \\
\bottomrule
\end{tabular}}

\caption{Performance comparison between our variant and previous state-of-the-art NAT models. All results reported are quoted from respective papers. \textbf{Iter.} is the number of decoding iterations, Adv. denotes adaptive, $^*$ denotes models trained with distillation data from Transformer-big. The Speedup is measured on WMT’14 En-DE test set with batch size 1. AT mE-nD refers to the AT model for re-scoring that has m encoder layers and n decoder layers.}
\label{tab:main-rst}
\end{table*}

\subsection{Main Results}
\label{main_results}
\paragraph{Generality of Candidate Soups}
Table \ref{tab:add2base} shows the performance improvement of our method for four base models on four translation tasks. Here, our number of candidate results is set to 5\footnote{If not specified below, the default number of candidate results is 5, and both NPD and Candidate Soups will be re-scored by AT teacher.}. And we use the Transformer-Base as the architecture of the AT model for re-scoring\footnote{If not specified below, the default architecture of AT model for re-scoring is Transformer-Base.}. The results show that for each NAT model and translation task, using Candidate Soups can achieve an average of \textbf{2.51-2.92} BLEU  higher than the base model. Even compared with the NPD, Candidate Soups improves BLEU by an average of \textbf{0.76-1.39} BLEU, which is a considerable improvement in machine translation task. Impressively, using the Candidate Soups on a strong baseline (GLAT \& DSLP) can achieve superior performance than the AT teacher. Furthermore, when AT models are used for re-scoring, they can perform parallel decoding as fast as training~\cite{gu2018non}. So the inference latency is only roughly doubled, which is still much faster than the AT model.

In conclusion, the above experimental results show that Candidate Soups is a general approach that can significantly improve translation quality while maintaining fast inference speed.

\paragraph{Comparing with the State of the Art}
To evaluate the best performance Candidate Soups can achieve, we compare our best variant (GLAT+DSLP+Candidate Soups) with previous state-of-the-art NAT models, including Iterative NAT and Fully NAT. As shown in Table \ref{tab:main-rst}, compared with the Iterative NAT, we produce a very competitive translation quality with approximately 2$\times$-4$\times$ faster inference speed. Compared with Fully NAT, our best variant is better than all existing models in two translation tasks (EN$\rightarrow$DE, RO$\rightarrow$EN) and is close to the current state-of-the-art performance in the remaining two translation tasks. More encouragingly, our approach even performed better than AT teacher on three translation tasks and achieved very comparable performance on the remaining one translation tasks, which extensively validated the effectiveness of Candidate Soups.

In addition, we also try to use two smaller AT models for re-scoring to accelerate the inference speed further. These two models have the same hyperparameters as Transformer-base, except for the number of layers of decoder and encoder. AT 4E-2D contains 4 encoder layers and 2 decoder layers, and AT 3E-1D contains 3 encoder layers and 1 decoder layer. Moreover, they were trained using the same distillation data as the NAT model. Surprisingly, even when the small AT models were used for re-scoring, our method maintained a similar performance to the previous model (AT 6E-6D), and its inference speed was \textbf{10.1}$\times$-\textbf{11.5}$\times$ that of the AT model. This result further proves that Candidate Soups can well balance the trade-off between translation quality and inference speed.

\subsection{Ablation Study and Analysis}
\label{analysis}
\paragraph{Influence of the Candidate Number}
\begin{figure}[t]
    \centering
    \includegraphics[width=0.41\textwidth]{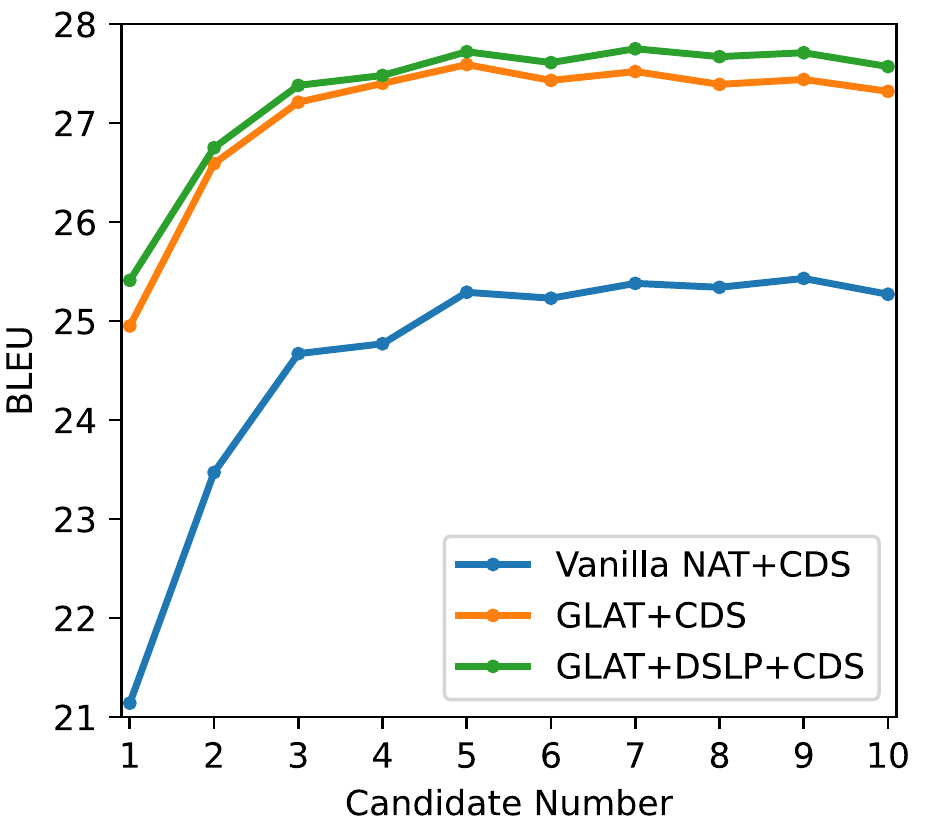}
    \caption{Translation quality under different number of candidate results on WMT'14 EN-DE.}
    \label{fig:Influence_number}
\end{figure}
In order to analyze the effect of the candidate translation number on the Candidate Soups, we conduct experiments with different candidate numbers.  Figure \ref{fig:Influence_number} shows the relationship between translation quality and the number of candidate results. Specifically, with the increased candidate numbers, the quality of the translation generally maintains a growth trend. Especially when the candidate results number is less than 4, the Candidate Soups performance is significantly improved when the number increases. However, when the number increases to a certain threshold, the quality of the translation begins to fluctuate, even showing a slight downward trend. We guess this is because when the number is larger than the threshold, the quality of the candidate translations may gradually decrease due to the gap between the predicted length and the actual length becoming larger. Furthermore, there may be duplication between the candidate results. Therefore, using 3-7 candidate translations is enough for Candidate Soups to significantly improve the quality of the final translation.

\paragraph{Influence of the Autoregressive Teacher}
\begin{figure}[t]
    \centering
    \includegraphics[width=0.44\textwidth]{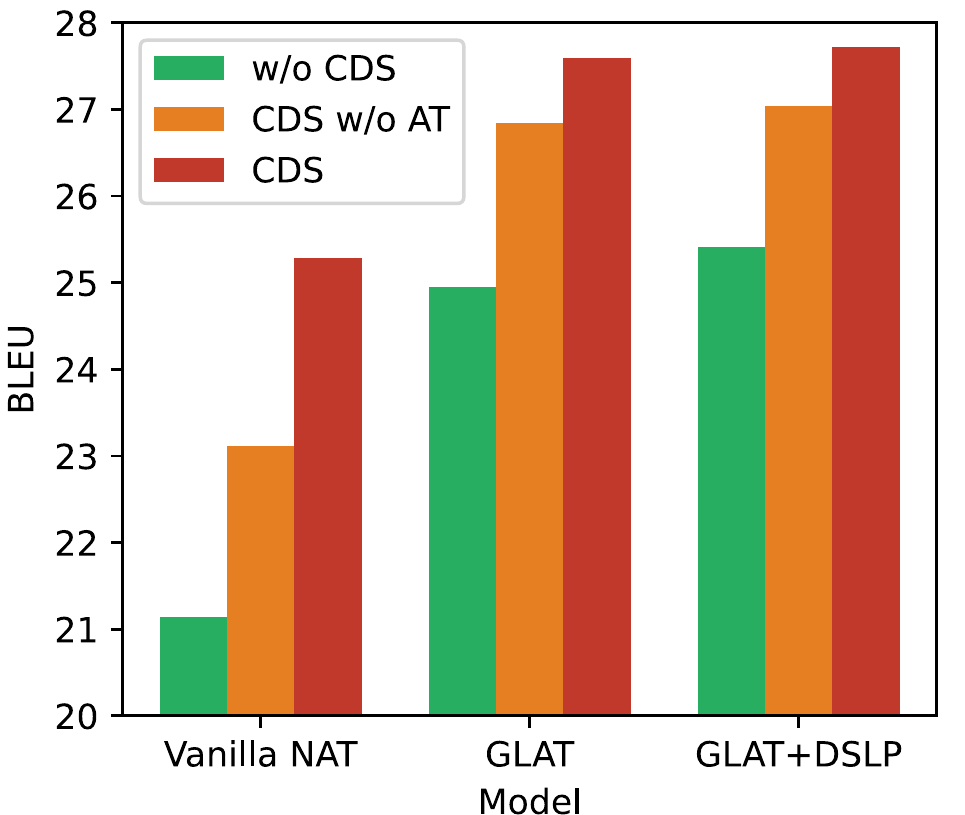}
    \caption{Performance of NAT model that with or without AT teacher re-scoring on WMT'14 EN-DE.}
    \label{fig:Influence_AT}
\end{figure}
To analyze the effect of whether using the AT model to re-score on our proposed method, we conducted experiments on the WMT’14 EN-DE dataset. Figure \ref{fig:Influence_AT} demonstrates that  Candidate Soups has a different dependence on AT model in different NAT models. For a model with weaker performance, such as Vanilla NAT, when AT model is not used for re-scoring, Candidate Soups' performance degrades significantly. However, for GLAT and GLAT+DSLP, which can produce high-quality translations, Candidate Soups can still increase approximately \textbf{1.89-1.63} BLEU even without re-scoring with the AT model. Notably, using Candidate Soups in this case hardly increases the inference time of the NAT model. Moreover, after using AT model for re-scoring, the effect of Candidate Soups can be further improved, which increases  \textbf{2.64} and \textbf{2.31} BLEU on the GLAT and GLAT+DSLP, respectively. In addition, the results in Table \ref{tab:main-rst} show that we can further improve the inference speed on the premise of guaranteeing translation quality by using a smaller AT model for re-scoring.

\paragraph{Influence of the Source Input Length}
\begin{figure}[t]
    \centering
    \includegraphics[width=0.43\textwidth]{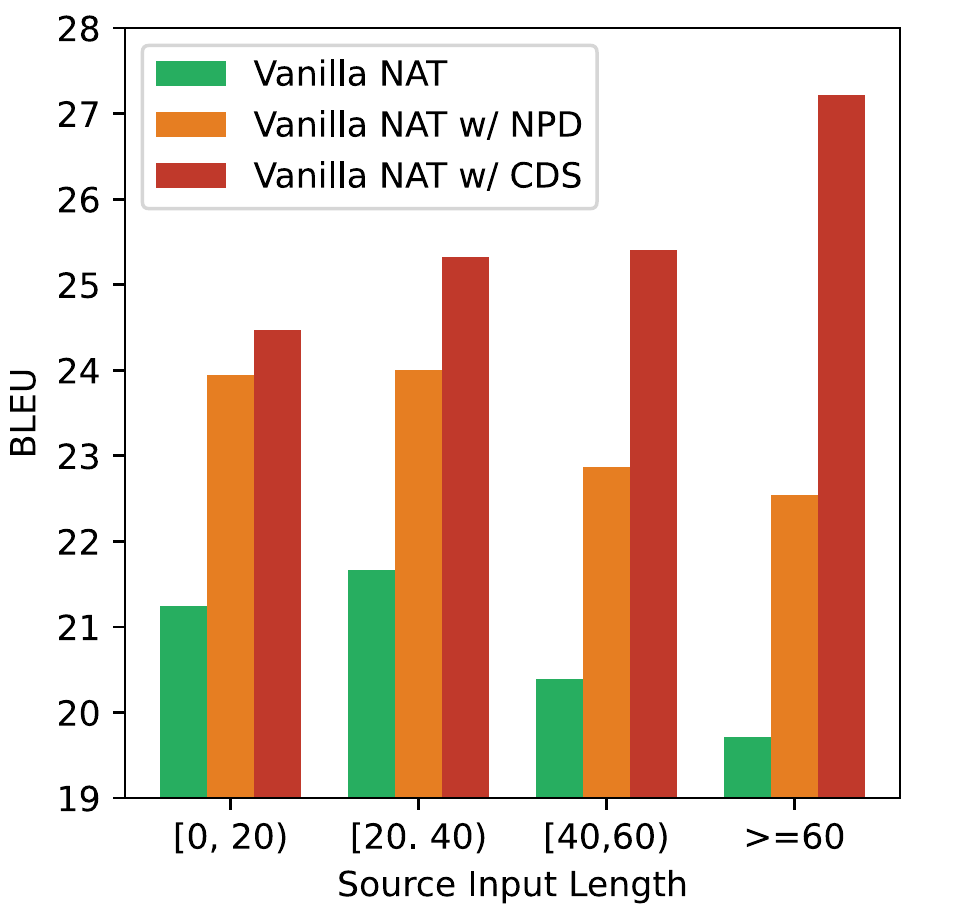}
    \caption{Performance under different source input
length on WMT'14 EN-DE.}
    \label{fig:Influence_Length}
\end{figure}
To analyze the influence of source sentence length on Candidate Soups' performance, we divide the source sentence after BPE into different intervals by length and calculate the BLEU score of each interval. The histogram of results is presented in Figure \ref{fig:Influence_Length}. It can be seen that the performance of Vanilla NAT degrades significantly as the length of the source sentence increases. Although NPD can improve the overall translation quality, the translation quality of long source sentences is still inferior. However, Candidate Soups can dramatically improve Vanilla NAT's performance and enables long sentences to achieve much higher BLEU than short. Impressively, the BLEU score of the source sentence length ranging from 40 to 60 increases by \textbf{5.01}, and the BLEU score of the source sentences longer than 60 increases by \textbf{7.51}.


We believe this is because the NAT model tends to generate more uncertain and diverse candidate results for longer source sentences. This feature enables Candidate Soups to obtain more useful information in the candidate results to generate higher-quality translations. These experimental results further verify the potential of the Candidate Soups in translating complex long sentences.  More experimental results and analyses are presented in the Appendix \ref{appendix:Analysis}.

\section{Conclusion}
In this paper, we propose “Candidate Soups,” which can discover and fuse valuable information from multiple candidate translations based on model uncertainty. This approach is general and can be applied to various NAT models. Extensive experimental results prove that the translation quality of the NAT model can be significantly improved by using Candidate Soups, especially for long sentences that are difficult to translate. And the trade-off between translation quality and inference speed is well controlled and balanced by Candidate Soups. Furthermore, our best variant can achieve better results on three translation tasks than the AT teacher while maintaining NAT's high-speed inference.

\section*{Limitations}


Although our proposed method can significantly improve the performance of non-autoregressive translation (NAT) models, it relies on trained autoregressive translation (AT) models to a certain extent. Not using the AT model for re-scoring can lead to poorer quality of translations generated by Candidate Soups, especially when using it for the poorer performing NAT model. Although using a small AT model is sufficient for Candidate Soups to achieve decent performance, it still results in a drop in inference speed and more GPU resources being used for translation. In addition, the performance of the AT model may limit the upper bound of the Candidate Soups' capability. Therefore, we will explore new methods that can be effective without AT re-score in the future.

\section*{Ethics Statement}
Our work has potentially positive implications for various non-autoregressive machine translation applications. It is a general method that can be applied to virtually all existing non-autoregressive translation models to improve their performance while maintaining their high inference speed. Our work can facilitate the implementation of non-autoregressive translation models in commercial companies and humanitarian translation services in the future and promote cultural exchanges between different languages and different races.

\section*{Acknowledgements}
This work was supported by NSFC grants (No. 62136002), National Key R\&D Program of China (2021YFC3340700) and Shanghai Trusted Industry Internet Software Collaborative Innovation Center. 

\bibliography{anthology,custom}
\bibliographystyle{acl_natbib}

\newpage
\appendix

\section{Background}
\label{appendix:background}
\subsection{Autoregressive Translation}
The autoregressive translation (AT) model achieves sort-of-the-art performance on multiple machine translation tasks~\cite{Song2019MASSMS,sun-etal-2020-multi}. Given a source sentence $X={\left(x_1, x_2, \ldots, x_n\right)}$ and the target sentence $Y={\left(y_1, y_2, \ldots, y_m\right)}$, the AT model decomposes the target distribution of translations according to the chain rule:
\begin{equation}
\begin{aligned}
    p_{\mathrm{AT}}(Y\mid X;\theta)=\prod_{t=1}^{m} p\left(y_{t}\mid y_{<t},X;\theta\right)
\end{aligned}
\label{eq:AT}
\end{equation}
where $y_{<t}$ denotes generated previous tokens before the $t^{th}$ position. During the training process, the AT model is trained via the teacher-forcing strategy that uses ground truth target tokens as previously decoded tokens so that the output of the decoder can be computed in parallel.

However, during inference, the AT model still needs to generate translations one by one from left to right until the token that represents the end \textbf{[EOS]} is generated. Although AT model has good performance, its autoregressive decoding method dramatically reduces the decoding speed and becomes the main bottleneck of its efficiency.

\subsection{Non-Autoregressive Translation}
To improve the inference speed, the non-autoregressive translation (NAT) model is proposed~\cite{gu2018non}, which removes the order dependency between target tokens and can generate target words simultaneously:
\begin{equation}
\begin{aligned}
    p_{\mathrm{NAT}}(Y\mid X;\theta)=\prod_{t=1}^{m} p\left(y_{t}\mid X;\theta\right)
\label{eq:NAT}
\end{aligned}
\end{equation}
where $m$ denotes the length of the target sentence. Generally, NAT models need to have the ability to predict the length because the entire sequence needs to be generated in parallel.
A common practice is to treat it as a classification task, using the information from the encoder's output to make predictions.

However, this superior decoding speed is achieved at the cost of significantly sacrificing translation quality. Because NAT is only conditioned on source-side information, but AT can obtain the strong target-side context information provided by the previously generated target tokens, there is always a gap in the performance of NAT compared with AT.

\subsection{Noisy Parallel Decoding}
Noisy parallel decoding (NPD)~\cite{gu2018non} is a stochastic search method that can draw samples from the length space and compute the best translation for each length as a candidate result. Then NPD selects the translation with the highest average log-probability as the final result:

\begin{equation}
\begin{aligned}
Y_{N\!P\!D}\!=\!\underset{Y^{m}}{\operatorname{argmax}}  \frac{1}{m}  \sum_{t=1}^{m} \log p_{N\!AT}\left(y_{t}^{m}\! \mid X\!; \theta\right)
\label{eq:NPD}
\end{aligned}
\end{equation}

where $Y^{m}$ is the translation predicted by the NAT model based on the length $m$. NPD also can use the AT model to identify the best translation:

\begin{equation}
\begin{aligned}
Y_{N\!P\!D}=\underset{Y^{m}}{\operatorname{argmax}}  \frac{1}{m}  \sum_{t=1}^{m} \log p_{AT}\left(y_{t}^{m}\! \mid y^{m}_{<t},X; \theta\right)
\label{eq:NPDwAT}
\end{aligned}
\nonumber 
\end{equation}

Notably, when an AT model is used for re-scoring, it can be decoded in parallel as it does at training time. Moreover, since all search samples can be computed independently, even with AT model for re-scoring, the latency of the NPD process is only doubled compared to computing a single translation.

\section{Additional Analysis Experiment}
\label{appendix:Analysis}

\subsection{Influence of the Knowledge Distillation}

Compared with the original data, the distillation data generated by the AT model has less noise and is more deterministic, which can effectively alleviate the multimodality problem of the NAT model. Therefore, almost all the existing NAT model adopts the method of Knowledge Distillation (KD) for training. However, generating distillation data tends to consume significant computing resources and time, and using distillation data to train NAT models may limit the translation capabilities of NAT models.

\begin{figure}[t]
    \centering
    \includegraphics[width=0.47\textwidth]{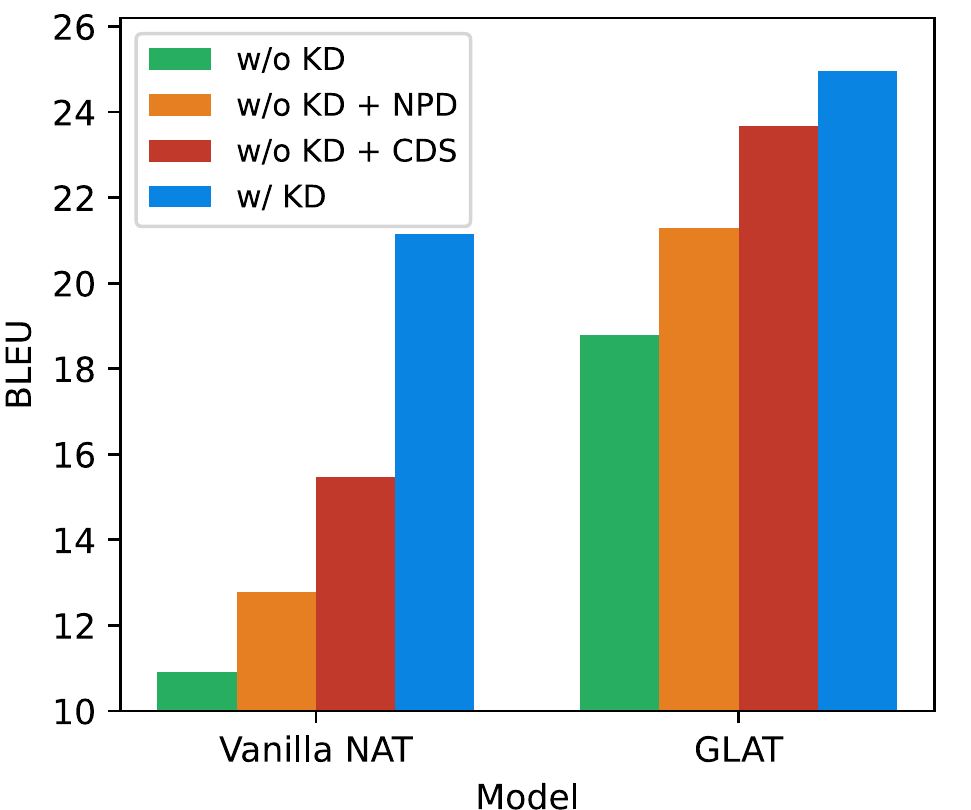}
    \caption{Performance of NAT model without Knowledge Distillation (KD) on WMT'14 EN-DE.}
    \label{fig:Influence_KD}
\end{figure}

In order to analyze whether our proposed method can still be effective in the scenarios where knowledge distillation is not used, we conducted experiments on the WMT’14 EN-DE dataset. Figure \ref{fig:Influence_KD} shows the performance of Candidate Soups on the NAT model that does not use knowledge distillation.  For Vanilla NAT and GLAT, the performance trained with raw data is significantly reduced compared to that trained with knowledge distillation. However, after using Candidate Soups, the BLEU of Vanilla NAT and GLAT respectively increased by \textbf{4.54} and \textbf{4.88}, which was only 5.68 and \textbf{1.28} lower than the performance with knowledge distillation. We believe that this significant performance improvement may be since NAT models without knowledge distillation may produce more diverse candidate translations, thus enabling Candidate Soups to fully play its role and obtain higher-quality translations from different candidate translations. The experimental results show that the Candidate Soups can significantly improve the performance of the NAT model without knowledge distillation, which proves the potential of the Candidate Soups in this scenario.

\subsection{Influence of introducing uncertainty methods}

In addition to introducing uncertainty through length, we propose two other methods for generating different candidate translations:
\begin{itemize}

  \item Use the prediction results of different decoder layers. DSLP~\cite{huang2021non} is a general method that can be applied to various NAT models, and it needs to predict the translations in each decoder layer. Therefore, Candidate Soups can be combined with DSLP, and any NAT model using DSLP can use Candidate Soups to fuse the results of different layers. In this experiment, we use the results generated by the last 5 layers of the decoder.
  
  \item Generate different translations by maintaining dropout during inference~\cite{Gal2016DropoutAA}. Even with the same input, the model can produce different outputs since dropout activates different neurons each time. In this experiment, the dropout probability at inference is set to 0.02.
  
\end{itemize}

\begin{table}[t]
\renewcommand{\arraystretch}{1.3}
\vspace{-.1cm}
\centering
\resizebox{.95\linewidth}{!}{
\Huge
\begin{tabular}{l|cc} 
\toprule[3pt]
\multirow{2}{*}{\textbf{Model}} & \multicolumn{2}{c}{\textbf{WMT’14}} \\ 
& EN-DE & DE-EN \\
\midrule
GLAT+DSLP & 25.41 & 29.28 \\
~~~~ w/ Candidate Soups(Length) & 27.72 & 30.98 \\
~~~~ w/ Candidate Soups(Layer) & 26.77 & 29.96 \\
~~~~ w/ Candidate Soups(Dropout) & 26.66 & 30.01 \\
\bottomrule[3pt]
\end{tabular}
}
\vspace{-.1cm}
\caption{Performance of Candidate Soups using different methods of introducing uncertainty. Length means introducing uncertainty with different lengths. Layer means using the prediction results of different decoder layers. Dropout means maintaining dropout during inference. The number of candidate translations is 5. }
\label{tab:inf_uncertainty}
\vspace{-.1cm}
\end{table}

Table \ref{tab:inf_uncertainty} shows the performance of Candidate Soup under three different ways of introducing uncertainty. The experimental results show that, compared with the other two methods, when uncertainty is introduced by length, Candidate Soups improves the translation quality more significantly. We speculate that this is because the length uncertainty can ensure that the generated translations are more diverse under the premise of high quality. 

However, for layer uncertainty, the quality of the translations produced by the first layer will be significantly lower than that of the last layer. These low-quality candidate translations are of little help to Candidate Soups and even affect the performance of Candidate Soups. For dropout uncertainty, the candidate translations generated will be affected by the dropout probability. On the one hand, if the dropout probability is set too high, it may reduce the overall quality of the candidate translations. On the other hand, the generated candidate translations will be less diverse if the dropout probability is low. So we further need to spend time searching for the optimal dropout probability setting for different NAT models and tasks. However, these two methods can still achieve about 1 BLEU improvement on the strong baseline (GLAT+DSLP), and their generalization ability is stronger than the length-based method. In addition, Candidate Soups can also be used as a new model ensemble method to enhance the final translation quality by using the output from multiple NAT models. We will discuss this in future work.

\end{document}